\documentclass[letterpaper, 10 pt, conference]{ieeeconf}  

\IEEEoverridecommandlockouts                              

\usepackage{color}
\usepackage{xcolor}
\usepackage{soul}
\usepackage{multirow}
\usepackage{booktabs}
\usepackage{url}
\usepackage{threeparttable}
\usepackage{cite}

\makeatletter
\let\NAT@parse\undefined
\makeatother

\usepackage[T1]{fontenc}
\usepackage{graphicx}
\usepackage{subfigure}

\usepackage{amsmath,amsfonts}
\usepackage[colorlinks,linkcolor=blue,anchorcolor=blue,citecolor=blue]{hyperref}
\usepackage{algorithmic}
\usepackage{algorithm}
\usepackage{array}
\usepackage{textcomp}
\usepackage{stfloats}

\usepackage{verbatim}

\usepackage{amssymb}
\usepackage{geometry}
\title{ \Large \bf
OTO Planner: An Efficient Only Travelling Once Exploration Planner for Complex and Unknown Environments
}

\author{Bo Zhou$^{*}$, Chuanzhao Lu, Yan Pan, and Fu Chen
\thanks{This work was supported by the National Natural Science Foundation (NNSF) of China under Grant 62073075. (Bo Zhou and Chuanzhao Lu contributed equally to this work.) (Corresponding author: Bo Zhou.)}
\thanks{The authors are with the School of Automation, Southeast University and the Key Laboratory of Measurement and Control of CSE, Ministry of Education, Nanjing 210096, P.R.China (emails: zhoubo@seu.edu.cn; lucz@seu.edu.cn; yanpan@seu.edu.cn; chenfu010728@gmail.com).
}
}

\begin{document}

\maketitle
\thispagestyle{empty}
\pagestyle{empty}

\begin{abstract}
Autonomous exploration in complex and cluttered environments is essential for various applications. However, there are many challenges due to the lack of global heuristic information. Existing 
exploration methods suffer from the repeated paths and considerable computational resource requirement in large-scale environments. To address the above issues, this letter proposes an efficient exploration planner that reduces repeated paths in complex environments, hence it is called ``Only Travelling Once Planner''.
OTO Planner includes fast frontier updating, viewpoint evaluation and viewpoint refinement. A selective frontier updating mechanism is designed, saving a large amount of computational resources. In addition, a novel viewpoint evaluation system is devised to reduce the repeated paths utilizing the enclosed sub-region detection. Besides, a viewpoint refinement approach is raised to concentrate the redundant viewpoints, leading to smoother paths. We conduct extensive simulation and real-world experiments to validate the proposed method. Compared to the state-of-the-art approach, the proposed method reduces the exploration time and movement distance by 10\%$\sim$20\% and improves the speed of frontier detection by 6$\sim$9 times.
\end{abstract}


\section{Introduction}

Autonomous exploration, which requires advanced intelligence and independence of the robot, has a broad spectrum of applications, including search and rescue, reconnaissance, online mapping, etc. Leveraging perception data from its own sensors, a robot can traverse and reconstruct the entire unknown region rapidly. During exploring, the robot is supposed to manage and make full use of a mass of data, so as to generate an optimal sequence of visiting, which is similar to the Travelling Salesman Problem(TSP) \cite{david2006traveling}. However, it is difficult to obtain the optimal order to traverse the unknown region due to the limited view from sensors and the lack of global heuristic information.

Current popular autonomous exploration methods \cite{dai2020fast,zhou2021fuel,cao2021exploring} utilize either RGB-D cameras or LiDARs (Light Detection and Ranging) as sensors. This letter focuses on the exploration using LiDAR, which is challenging due to the sparsity, high complexity of point cloud and the lack of color information of the objects \cite{cao2023representation}.
\begin{figure}[!t]
\centering
\subfigure[]{
\includegraphics[width=0.22\textwidth]{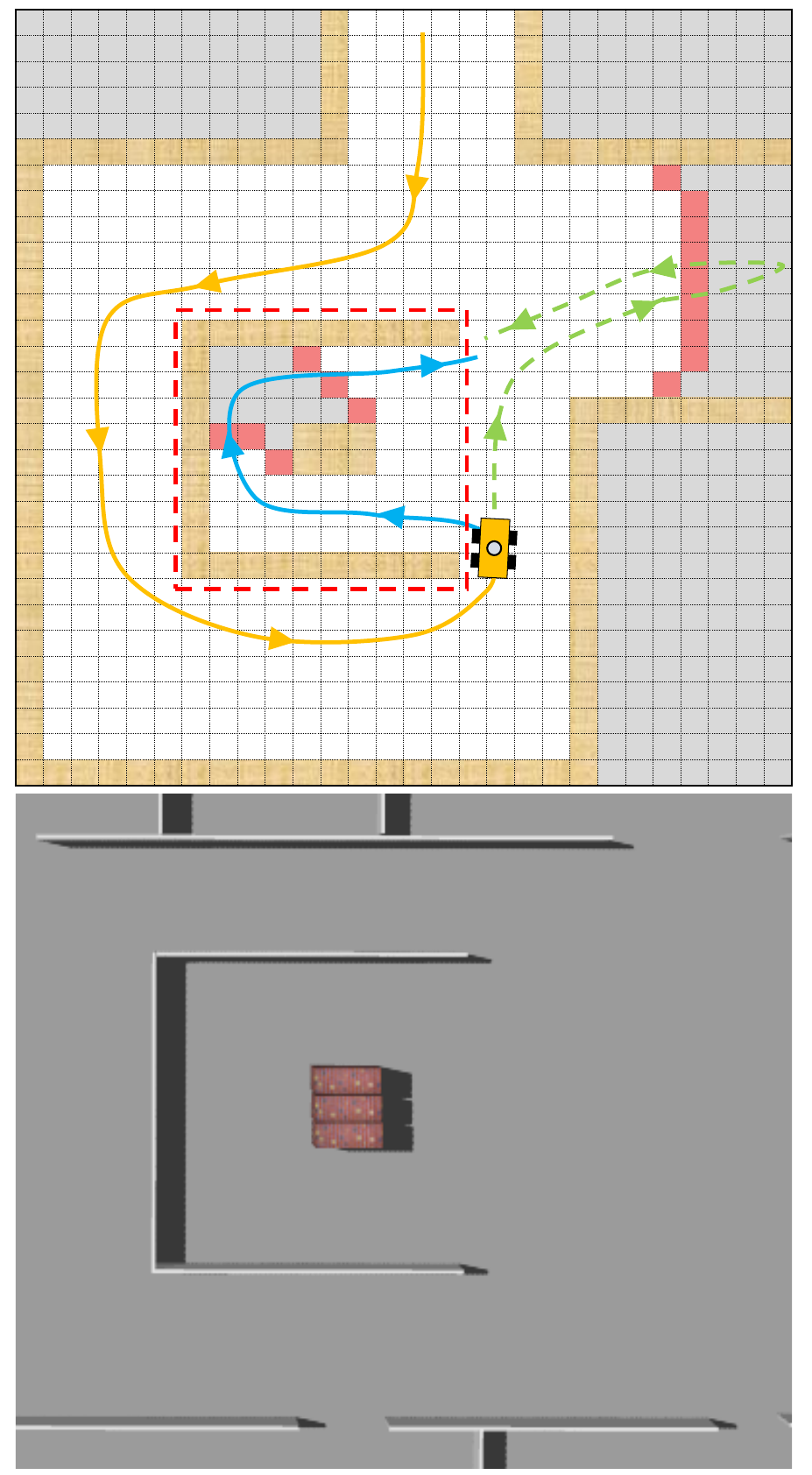}
}
\subfigure[]{
\includegraphics[width=0.22\textwidth]{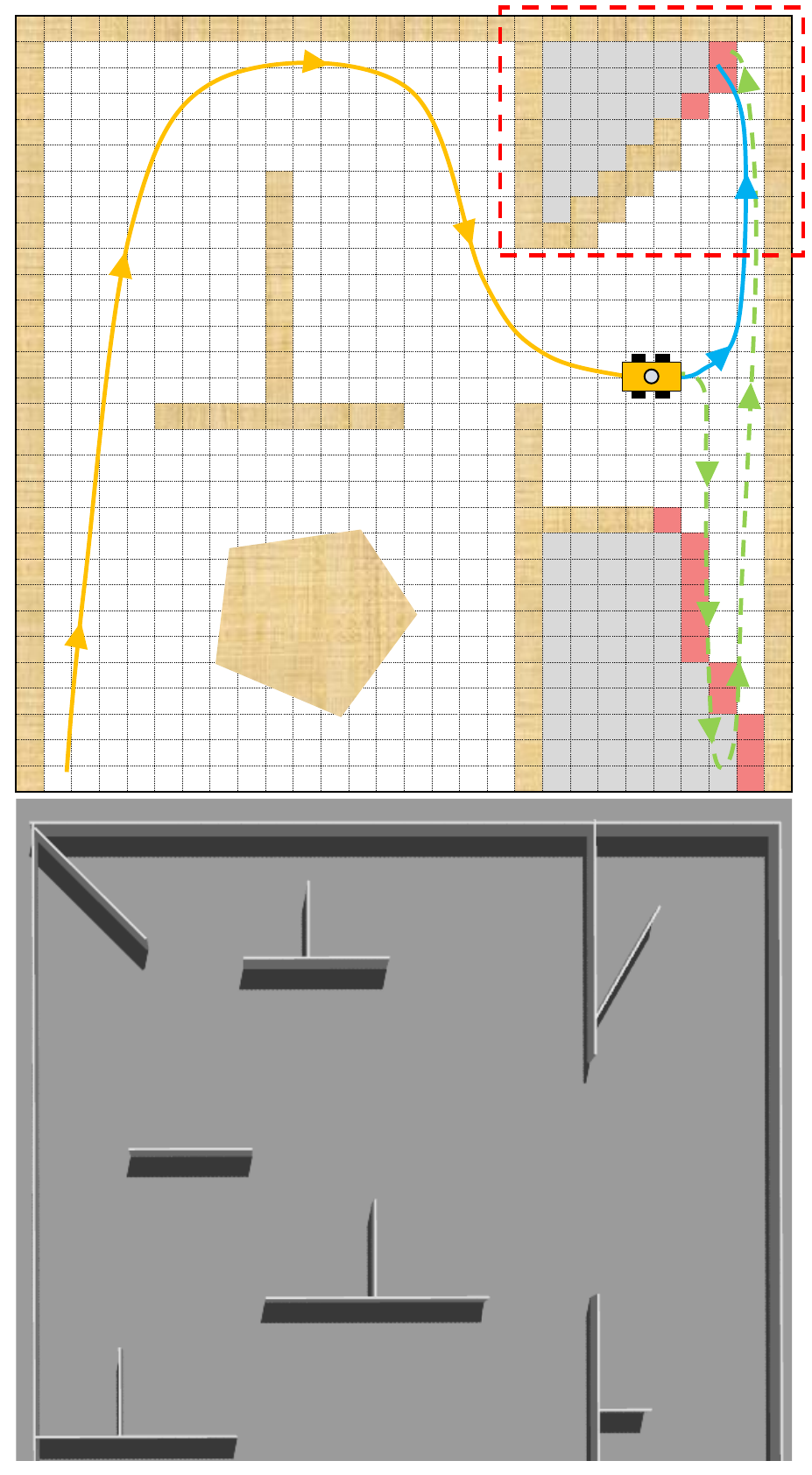}
}
\caption{Complex scenes that may cause repeated paths. The upper figures are diagrams and the lower figures are actual scenes in simulation. Gray grids: unknown spaces. Red grids: frontiers. Orange path: exploration path. Greedy strategies are prone to visit frontiers with maximum information gain, like green path (dashed line), which drive the robot back to visit the overlooked small regions. The red box shows the enclosed sub-region. If the robot is located near the enclosed sub-region, it prefers to explore unknown spaces in it, like blue path (solid line), which reduces repeated paths significantly.}
\label{illustrate}
\end{figure}

It should be noted that there are several crucial issues associated with efficient exploration for complex environments and this letter primarily discusses and addresses two of them. The first issue lies in the prevalent use of greedy strategies in existing methods when generating visiting sequence, which typically prefer to visit the region with the maximum information gain or closest to the robot\cite{ericson2021understanding}. In practical environments, local optima may not regularly lead to the optimal global efficiency since there are various small regions in cluttered environments. If these little regions are overlooked and the robot is navigated to a region far apart but with greater information gain, it is necessary for the robot to return to them, which inevitably results in repeated exploration paths, as shown in Fig. \ref{illustrate}. Consequently, developing a system capable of a comprehensive evaluation of viewpoints is essential. Several approaches \cite{yu2023echo, zhao2023autonomous} presuppose pre-defined scene boundaries to guide the robot in prioritizing exploration near these edges. Nevertheless, the scene to be explored is not always regular and the precise boundary is not always readily available in actual applications. The second issue pertains to the computationally intensive task of frontier detection, which depicts the boundary between known and unknown regions.
In large-scale environments, performing frontier detection in the overall map can be exceedingly time-consuming\cite{sun2022ada}, leading to a severe delay in environmental information processing.

Motivated by the aforementioned cases, based on TARE, this letter proposes an efficient \textbf{O}nly \textbf{T}ravelling \textbf{O}nce exploration planner for complex environments called \textbf{OTO Planner}, which can detect frontiers efficiently, reduce repeated paths without any prior information and generate smooth paths. 
The main contributions of this letter are summarized as follows:

\begin{itemize}
\item An efficient exploration framework called OTO Planner is raised, evaluating and refining viewpoints comprehensively to effectively address the local optima issue of commonly used greedy strategies. The framework considers coverage, changes in direction, travelling distance and the presence of enclosed sub-regions, reducing repeated paths significantly without prior knowledge of the scene.
\item  A fast LiDAR-based frontier updating mechanism is designed. Specifically, frontier updating only happens in areas that are newly perceived and unknown in the previous moment (only 20\%$\sim$30\% of the overall map), which saves considerable computational resources.

\item  Extensive validations in simulation and real-world environments are conducted. Our method is compared with classic\cite{bircher2016receding,dang2020graph} and state-of-the-art\cite{cao2021exploring} methods and the results show that our method reduces the exploration time and movement distance by 10\%$\sim$20\% compared to TARE. Meanwhile, the speed of frontier detection is 6$\sim$9 times faster than TARE.

\end{itemize}
The source code will be released to benefit community\footnote[1]{https://github.com/Luchuanzhao/OTO-Planner}.



\section{Related Work}
Autonomous exploration of robots has attracted broad attention in recent decades with the development of perception technology and computer science. The main task of robot exploration is to 
find the shortest path to traverse the entire unknown region quickly and completely. According to the exploration strategy, the previous methods for robot exploration can be roughly divided into three categories: sampling-based methods, frontier-based methods, and learning-based methods \cite{georgakis2022uncertainty,chaplot2020learning,cao2023ariadne}, which have recently emerged with the advancement in computational power. This letter only discusses two mature and widely used methods: sampling-based methods and frontier-based methods.

\subsection{Sampling-based Methods}

Sampling-based methods\cite{sun2022ada,bircher2016receding,dang2020graph,schmid2020efficient,dharmadhikari2020motion,xu2021autonomous} sample points in the free region of the environment, and then determine the priority of these points based on their positions through utility or cost functions. The higher priority points are selected as viewpoints to be visited in order.

The typical sampling-based exploration method RH-NBVP \cite{bircher2016receding} employs Rapidly-exploring Random Trees(RRT) \cite{lavalle1998rapidly} to determine the ``next-best-view''. Applying receding horizon planning strategy, the current position is used as the root node to reconstruct RRT after the robot visits the node with the maximum information gain. This exploration method is probabilistically complete, but there are still drawbacks such as repeated paths and high computation complexity. On this basis, \cite{witting2018history} utilizes historical visited viewpoints as potential RRT seeds, which can quickly lead the robot to next informative region when it falls into a dead-end situation. AEP \cite{selin2019efficient} combines frontier-based exploration and NBVP, where NBVP is employed as local exploration strategy and frontier information is mixed for global exploration to avoid getting stuck locally. Rapidly-Random Graph \cite{karaman2011sampling} is used in GBP \cite{dang2020graph} to solve the ``next-best-view'' problem. In this planning architecture, local planner constructs a dense RRG to evaluate robot paths with highest information gain and simultaneously avoid collision. While exploring large and complex scene, sparse RRG is engaged in global planner to get the robot out of a dead-end and navigate to next informative sub-region. GBP requires a large computational burden due to computing information gain of all nodes in local and global RRG. DSVP \cite{zhu2021dsvp} raises biased sampling, dynamically maintaining and expanding a RRT during the exploration and guiding the RRT with both local and global frontiers, with no need to compute the information gain of all nodes.



\begin{figure}[!t]
\centering
\includegraphics[width=0.5\textwidth]{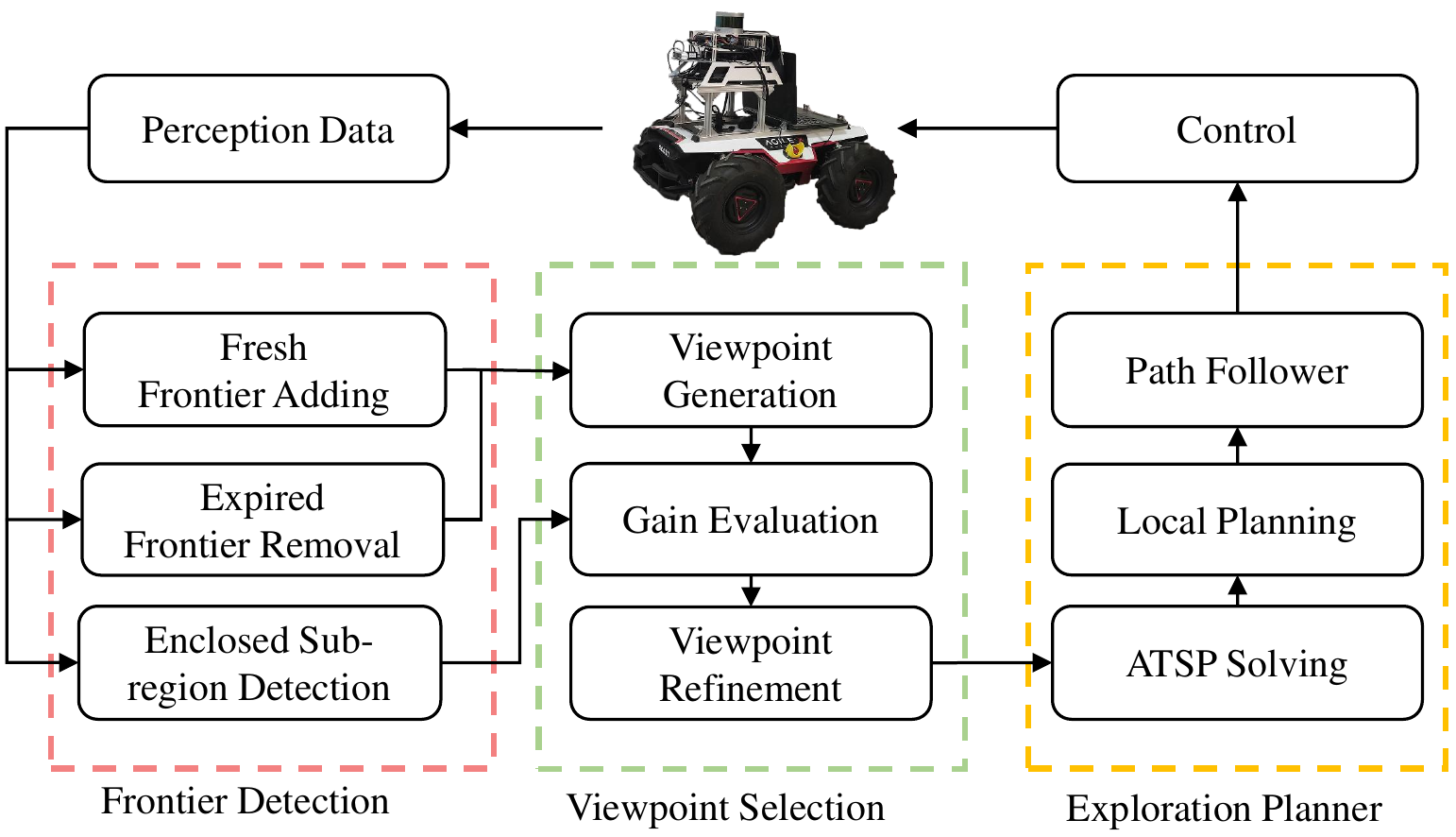}
\caption{An overview of proposed exploration approach. Our contributions are shown in frontier detection moudle and viewpoint selection moudle.}
\label{flowchart}
\end{figure}

\subsection{Frontier-based Methods}

Frontier-based methods \cite{zhou2021fuel,yu2023echo,huang2023fael,cao2021exploring,deng2020robotic,schmid2021unified} first find frontiers that are boundaries between known and unknown regions. Then, the frontier with the highest information gain will be selected as the next target. The original frontier-based method \cite{yamauchi1997frontier} always selects the closest frontier as the robot's next goal position until there is no unknown region. On the basis of this seminal work, numerous researchers make improvements and apply them to various fields. \cite{cieslewski2017rapid} presents an evaluation method to select the frontier that minimizes the velocity speed change, which is beneficial for UAV to maintain a high flight speed. FUEL \cite{zhou2021fuel} proposes frontier information structure (FIS) and clusters frontiers. After computing information gain of frontiers, it finds a visiting sequence by solving Asymmetric Traveling Sales Problem (ATSP) and plans B-spline trajectories. ECHO \cite{yu2023echo}  synthesizes boundary cost and environment structure cost based on FUEL, which can significantly improve exploration speed. However, accurate boundary information is often difficult to obtain in practical environments. TARE \cite{cao2021exploring} divides the whole region into some subspaces. In local planning, viewpoints are refined and traversed by solving TSP, while in global planning, subspaces are utilized in TSP to accelerate the exploration process. FAEL \cite{huang2023fael} raises a fast autonomous exploration framework, which employs the UFOMap \cite{duberg2020ufomap} for fast environmental information preprocessing and designs an path optimization formulation.

Inspired by TARE, this letter proposes a fast and low computational resources required exploration approach, which updates frontiers selectively and leverages the environment structure information.

\section{Methodology}

\begin{figure}[!t]
\centering
\includegraphics[width=0.47\textwidth]{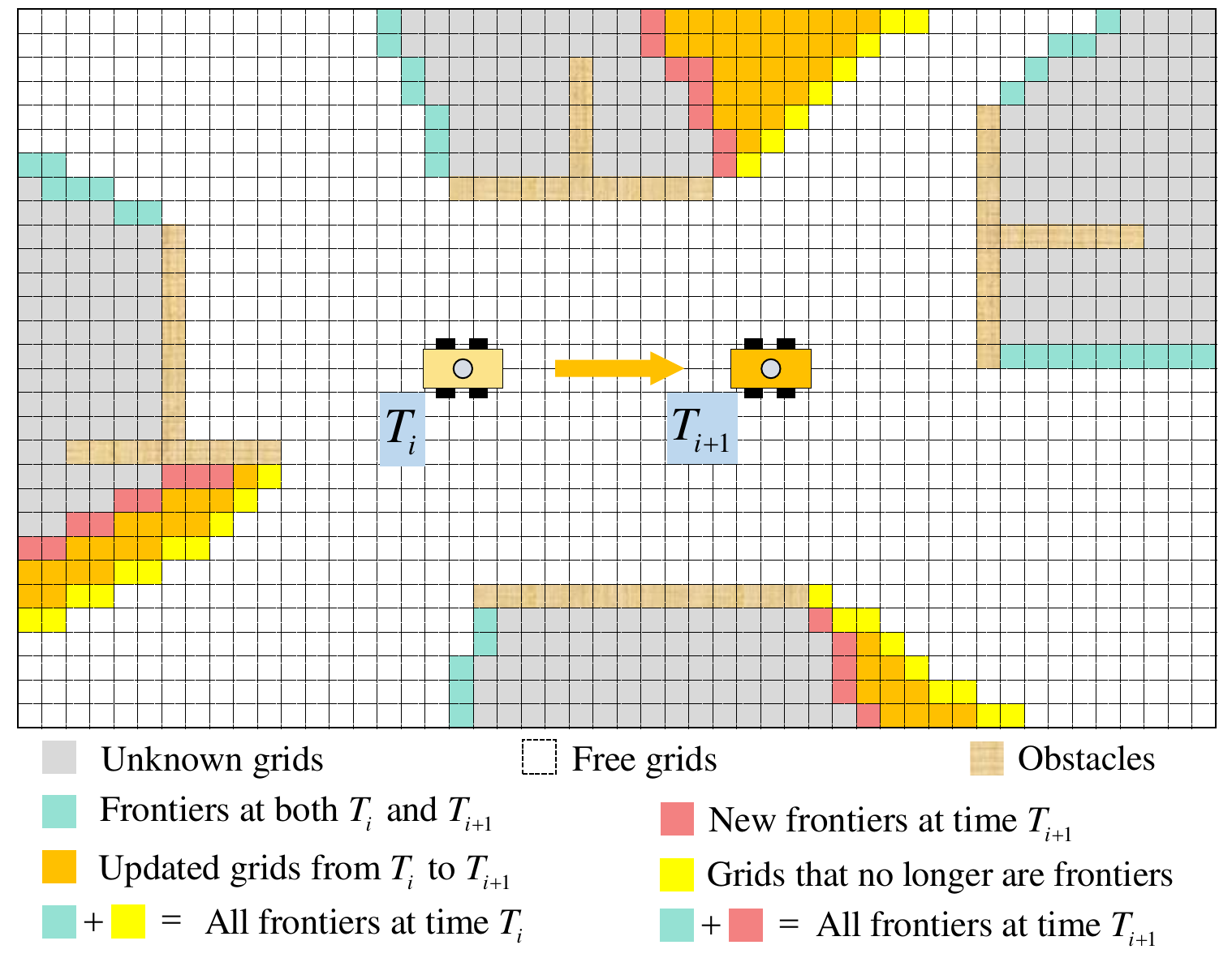}
\caption{Selective frontier updating. As robot moves from time $T_i$ to $T_{i+1}$, some unknown grids are newly perceived, like orange grids in the figure. All frontiers at time $T_i$ are checked whether they still meet the features of frontiers. In the meantime, fresh frontiers are detected in the updated (orange) grids.}
\label{frontier}
\end{figure}

The overall process of proposed exploration planner is shown in Fig. \ref{flowchart}. The perception data is first transmitted into frontier detection module and frontiers are updated selectively. Subsequently, viewpoints are generated in free spaces and the cost of each viewpoint is evaluated. After enclosed sub-region is detected, costs of viewpoints will be reevaluated according to whether they are in enclosed sub-region. Then, viewpoint refinement will be executed to concentrate the viewpoints according to their position, which leads to less tortuous path. Finally, exploration sequence is acquired by solving ATSP, and local planner is employed to optimize the path and drive the robot to reach the next goal.

\subsection{Selective Frontier Updating}
In most exploration approaches, frontiers are exploited in viewpoint generation and information gain evaluation. Therefore, frontier identification is an essential part that determines whether the exploration planner can work efficiently. In previous work \cite{cao2021exploring}, all grids in the partially constructed map need to be traversed at a fixed frequency. In large-scale environments, frontier identification requires considerable computational resources. To achieve fast frontier identification, we present a frontier detector that updates frontiers only in newly perceived areas.

As shown in Fig. \ref{frontier}, from $T_i$ to $T_{i+1}$, the sensor on the robot perceives new regions. Fresh frontiers and expired frontiers are only updated in the newly perceived areas instead of in the overall map. A grid is considered to be frontier if its state is free and there is at least one unknown grid adjacent to it. Define $F_{T_i}$ as the frontiers at time $T_i$, $F_{T_{i+1}}$ as the frontiers at time $T_{i+1}$ and $G_i^{i+1}$ as the newly perceived grids from $T_i$ to $T_{i+1}$. To reduce the computational resources for frontier updating, fresh frontiers are only detected in $G_i^{i+1}$ because new frontiers only appear in newly updated regions. In addition, $F_{T_i}$ may no longer be frontiers owing to its neighboring grids being updated. In Fig. \ref{frontier}, the yellow grids are part of $F_{T_i}$ at $T_i$, while they are removed at $T_{i+1}$ because the state of grids surrounding the yellow grids are updated as known.

\begin{figure}[!t]
\centering
\includegraphics[width=0.47\textwidth]{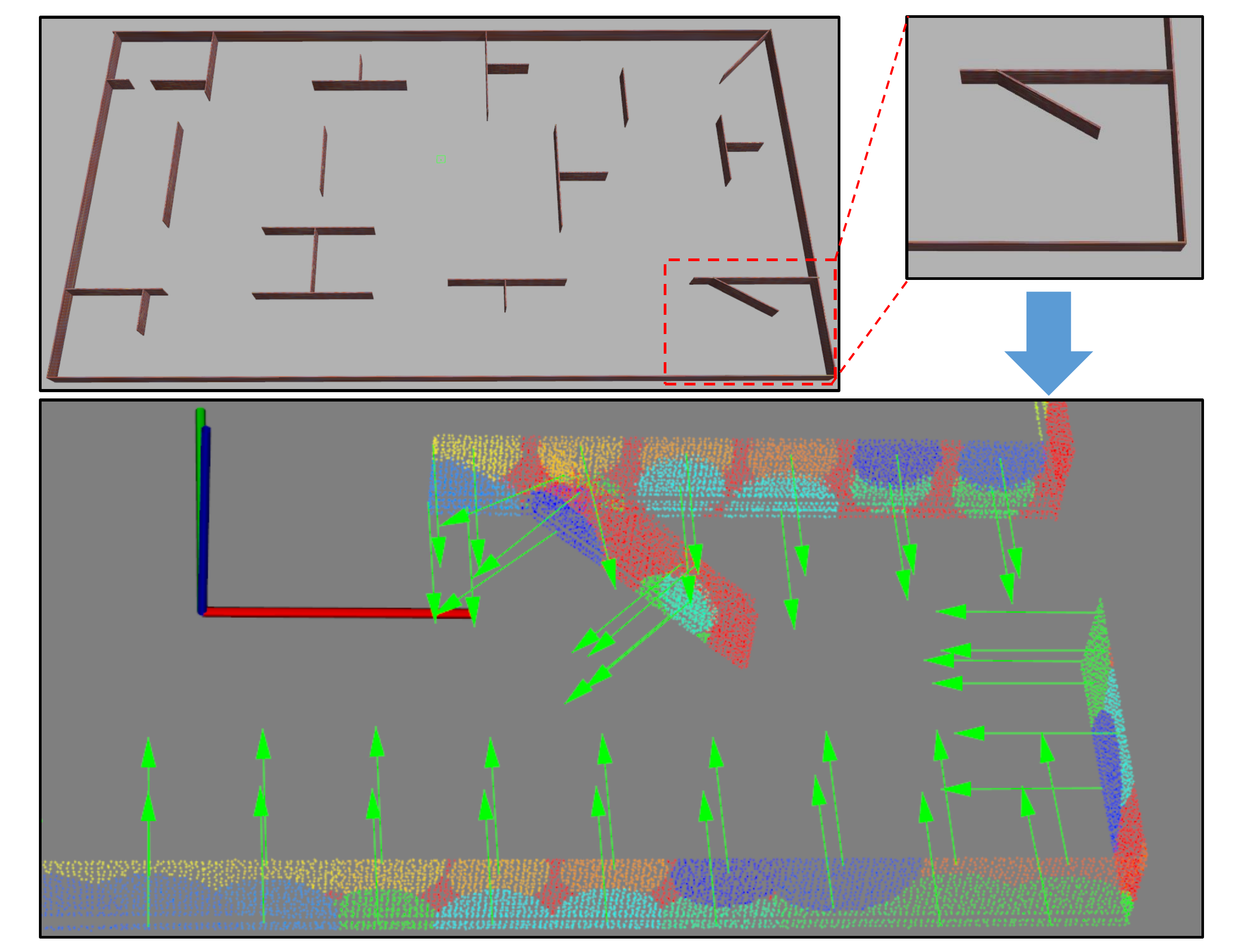}
\caption{Enclosed sub-region detection. \textbf{Upper}: the whole scene for the robot to explore and a portion of the scene that is centered and extracted around the robot. \textbf{Lower}: the process of detecting enclosed sub-region. Point clouds of different colors represent regions within distinct neighborhoods. The green arrow represents the normal vector of a point cloud neighborhood.}
\label{enclosed}
\end{figure}


\subsection{Viewpoint Evaluation}
\label{III-B}

For frontier-based exploration methods, designing an effective sequence to visit viewpoints is a significant step. Therefore, viewpoint evaluation plays an important role in determining the priority of each viewpoint, which subsequently affects the efficiency of the exploration process. 

In actual applications, accurate boundary may not always be available. As a result, heuristic exploration approaches \cite{yu2023echo, zhao2023autonomous} will not work very well. In cluttered environments, greedy strategies are prone to select viewpoints with maximum information gain, which will lead to a problem that small regions may be overlooked in local planning and need to be visited after other regions being explored completely. If there are so many small regions being overlooked, robot has to come back to visit these regions, which will lead to repeated paths and a considerable time consumption.

\renewcommand{\algorithmicrequire}{\textbf{Input:}}
\renewcommand{\algorithmicensure}{\textbf{Output:}}

\begin{algorithm}[!t]
\caption{Enclosed Sub-region Detection.}
\begin{algorithmic}[1]
\REQUIRE point cloud map $\mathcal{M}_{cur}$ around the robot
\ENSURE boundary of enclosed sub-region $B_{enclosed}$
\STATE $\textbf{Initial}$ the number of normal vectors in each quadrant\\ \hspace{0.5cm} $cnt_j$=$0$, state of quadrant $Q_j$=free $\textbf{for}$ $j = 1,...,4$
\STATE $B_{enclosed}=\varnothing$
\STATE $\mathcal{M} = \textbf{Preprocess}(\mathcal{M}_{cur})$
\STATE sampled points $P \gets \textbf{UniformlySample}(\mathcal{M}, r)$
\STATE \textbf{for each} $p_i$ in sampled points $P$ \textbf{do}
\STATE \hspace{0.5cm} nearest $n$ points $P_{local} \gets$\\ \hspace{0.7cm} $\textbf{KNearestNeighbor}(p_i, \lambda)$
\STATE \hspace{0.5cm} normal vector $\vec{n}_i$ = $\textbf{CalNormalVector}(P_{local})$
\STATE \hspace{0.5cm} angle $\alpha_i$ = $\textbf{GetAngle}(\vec{n}_i, Plane_{XOY})$
\STATE \hspace{0.5cm} $\textbf{if}$ $ \alpha_i > \alpha_{max} $ $\textbf{then}$
\STATE \hspace{1cm} $\textbf{continue}$
\STATE \hspace{0.5cm} $\textbf{end if}$
\STATE \hspace{0.5cm} $\textbf{for}$ j := 1 to 4 $\textbf{do}$
\STATE \hspace{1cm} $\textbf{if}$ $\vec{n}_i$ in the $j_{th}$ quadrant $\textbf{then}$
\STATE \hspace{1.5cm} $cnt_{j}$++
\STATE \hspace{1cm} $\textbf{end if}$
\STATE \hspace{1cm} $\textbf{if}$ $cnt_j > \lambda$ $\textbf{then}$
\STATE \hspace{1.5cm} $Q_j$ = occupied
\STATE \hspace{1cm} $\textbf{end if}$
\STATE \hspace{0.5cm} $\textbf{end for}$
\STATE $\textbf{end for}$
\STATE m = $\textbf{GetNumberOfOccupiedQuadrant}$($Q_j$)
\STATE $\textbf{if}$ m $\geq$ 3 $\textbf{then}$
\STATE \hspace{0.5cm} $B_{enclosed} \gets \textbf{CalBoundingBox}(\mathcal{M})$
\STATE \textbf{return}  $B_{enclosed}$
\end{algorithmic}
\label{enclosed_detection}
\end{algorithm}

To address this issue, this letter proposes a concept of enclosed sub-region. If the robot is located near a enclosed region, it prefers to visit the viewpoints in these regions. Algorithm. \ref{enclosed_detection} illustrates the process of enclosed sub-region detection. Only the enclosed sub-regions around the robot are to be explored, because robot may not explore the regions far apart due to large path length cost. We will only detect point clouds within a cubic region centered on the robot. Since the environments that most robots need to explore can be simplified into two-dimensional spaces, the enclosed sub-region detection in this letter does not take into account features along the Z-axis. Before detection, the point cloud map around the robot is preprocessed by removing the ground points. For efficient nearest neighbors search, K-Dimensional Tree (K-D Tree) \cite{bentley1975multidimensional} is employed.
We then uniformly sample points at certain interval $r$ on the processed point cloud map and denote them as $p_i$. As shown in Fig. \ref{enclosed}, centered at $p_i$, we perform nearest neighbors search in kd-tree to obtain $n$ points. After that, normal vectors are calculated for every $n$ points centered at $p_i$, which are denoted as $\vec{n}_i$. 

We define a coordinate system with its origin at the center of the local point cloud map (i.e. the position of the robot), where the x-axis and y-axis are aligned with the x-axis and y-axis of the global map, respectively.

To filter out the regions that are enclosed by multiple surfaces, within the local coordinate system, we adjust all normal vectors to orient them towards the local coordinate origin and determine which quadrant the normal vectors are located in. In the meantime, if the angle between $\vec{n}_i$ and the XOY-plane exceeds $\alpha_{max}$, $\vec{n}_i$ is discarded to eliminate interference in the direction of the Z-axis. To prevent interference from small objects, a quadrant is considered occupied when it contains at least $\lambda$ normal vectors. Subsequently, if there are at least 3 quadrants that are occupied, the region surrounded by these surfaces is probably enclosed sub-region. Specially, the boundary of enclosed sub-region is supposed to be calculated carefully.
In order to illustrate it more concisely, we define the bounding box of enclosed sub-region as $B_{enclosed}$. 
For all surfaces, to simplify the process without significantly sacrificing accuracy, we employ planes to fit the points within each quadrant.
In this paper, RANSAC \cite{fischler1981random} is employed in each quadrant to fit the planes and determine the boundary limits of the bounding box. Note that in the RANSAC algorithm, only the uniformly sampled points $p_i$ are utilized, which requires minimal computational resources.

For each fitted plane denoted as $ Plane_k $, we determine its maximum and minimum x and y coordinates, which are respectively denoted as $ {x_k}_{max} $, $ {y_k}_{max} $, $ {x_k}_{min} $, and $ {y_k}_{min} $. Practically, we find that taking the boundaries of the planes within each quadrant is effective to represent the boundaries of the enclosed region, which can be defined as:

\begin{equation}
\left\{
\begin{array}{lr}
{B_x}_{max} = max({x_k}_{max})  \\
{B_x}_{min} = min({x_k}_{min})  \\
{B_y}_{max} = max({y_k}_{max})  \\
{B_y}_{min} = min({y_k}_{min}) 
\end{array}
\right.
\end{equation}
where $k \in \left\{1,...,4 \right\}$, $B_x$ and $B_y$ are the boundaries of the enclosed sub-region.

In order to enable viewpoints in enclosed sub-regions to be visited with priority, we calculate the sub-region cost $c_r(i)$ for each viewpoint $v_i$:

\begin{equation}
c_r(i)=
\begin{cases}
0 & \text{if } v_i \text{ is located in } B_{enclosed} \\
dis(v_i, C_{near}) & \text{otherwise}
\end{cases}
\end{equation}
where $C_{near}$ represents the center of nearest $B_{enclosed}$ from $v_i$ and $dis(v_i, C_{near})$ represents the Euclidean distance from $v_i$ to $C_{near}$. 

Distance from the robot's current position to the viewpoint is another critical factor in assessing the priority of the viewpoint. As is introduced in \cite{cao2021exploring}, we employ the path length obtained from the A* algorithm as the cost:

\begin{equation}
c_l(i)=AStarPathLength(p_{cur}, v_i)
\end{equation}
where $p_{cur}$ denotes the current position of the robot.

Due to the significant time consumption and consequent reduction in exploration efficiency caused by the robot's back-and-forth maneuvers, we take into account the change in the robot's direction, with the cost calculation as follows:

\begin{equation}
c_d(i)=\arccos{\frac{(p_{cur}-p_{v_i}) \cdot \textbf{v}_{cur}}{\Vert (p_{cur}-p_{v_i}) \Vert \cdot \Vert \textbf{v}_{cur} \Vert}}
\end{equation}
where $p_{cur}$ is the current position of the robot, $p_{v_i}$ is the position of the $i_{th}$ viewpoint and $\textbf{v}_{cur}$ is the current velocity of the robot.

Finally, the utility of each viewpoint can be calculated as:
\begin{equation}
\textbf{U}(v_i)=\textbf{G}(v_i) \cdot e^{-c_{total}}
\end{equation}
\begin{equation}
c_{total}=w_r \cdot c_r + w_l \cdot c_l + w_d \cdot c_d
\label{equ6}
\end{equation}
where $\textbf{U}(v_i)$ is the utility of $v_i$, $\textbf{G}(v_i)$ is the information gain of $v_i$, $c_{total}$ is the total cost and $w_r$, $w_l$, $w_d$ are the corresponding weights of the three cost terms respectively.

Inspired by \cite{zhou2021fuel}, the ATSP is subsequently utilized to generate a sequence for traversing all viewpoints, both local and global.

\renewcommand{\algorithmicrequire}{\textbf{Input:}}
\renewcommand{\algorithmicensure}{\textbf{Output:}}

\begin{algorithm}[!t]
\caption{Viewpoint Refinement Strategy.}
\begin{algorithmic}[1]
\REQUIRE all local viewpoints $V_{ini}$ at the current time, distance threshold $D_{thr}$
\ENSURE viewpoints $V_{ref}$ after being refined
\STATE $V_{ref}= V_{ini}$
\STATE \textbf{for each} $v_i$ in current viewpoints $V_{ref}$ \textbf{do}
\STATE \hspace{0.5cm} $V_{near} = \textbf{RadiusNearestNeighbor}(v_i, D_{thr})$
\STATE \hspace{0.5cm} $V_{cluster} = \textbf{GetUnobstructedVPs}(V_{near})$
\STATE \hspace{0.5cm} $\textbf{if}$ $V_{cluster} \neq \varnothing$ $\textbf{then}$
\STATE \hspace{1cm} $v_{avg}= \textbf{CalCentroid}(V_{cluster})$
\STATE \hspace{1cm} $V_{ref}$.erase($V_{cluster}$)
\STATE \hspace{1cm} $V_{ref}$.push($v_{avg}$)
\STATE \hspace{0.5cm} $\textbf{end if}$
\STATE $\textbf{end for}$
\STATE \textbf{return}  $V_{ref}$
\end{algorithmic}
\label{viewpoint_refinement}
\end{algorithm}

\subsection{Viewpoint Refinement}
\label{III-C}

During the exploration process, viewpoints are generated randomly in free regions and are subsequently filtered based on their potential to cover frontiers, which results in a high degree of randomness in the distribution of viewpoints and inefficiency of TSP. As the robot explores unknown scene, it has to visit each viewpoint. However, an overly random distribution may necessitate a convoluted traversal path for the robot, which can substantially impair the efficiency of the exploration process.

To mitigate the issue of path convolution caused by the random distribution of viewpoints, this paper proposes a novel viewpoint refinement approach that can lead to a smoother trajectory. For each local viewpoint $v_i$, if the distance to another viewpoint $v_j$ is less than the threshold $D_{thr}$, $v_j$ is recorded in the set $V_{near}$. If these viewpoints, which are close enough to each other, are unobstructed by any obstacles, they are denoted as a set $V_{cluster}$. Instead of remaining every individual viewpoint, we replace the positions of these clustered viewpoints with their geometric centroid, and their information gain with the average value of them, denoted as $v_{avg}$. This treatment of viewpoints not only reduces the number of way points the robot needs to traverse but also ensures that the original frontiers are not missed. As the robot progressively visits these refined viewpoints, some frontiers may already be covered, eliminating the need to access certain viewpoints directly and thus avoiding convoluted routes. Although this approach might potentially overlook some minor frontiers, the enclosed sub-region detection algorithm in Section \ref{III-B} will drive the robot to regenerate viewpoints to cover such frontiers if necessary. More details can be seen in Algorithm \ref{viewpoint_refinement}.

\begin{figure}[!t]
\centering
\subfigure[]{
\includegraphics[width=0.21\textwidth]{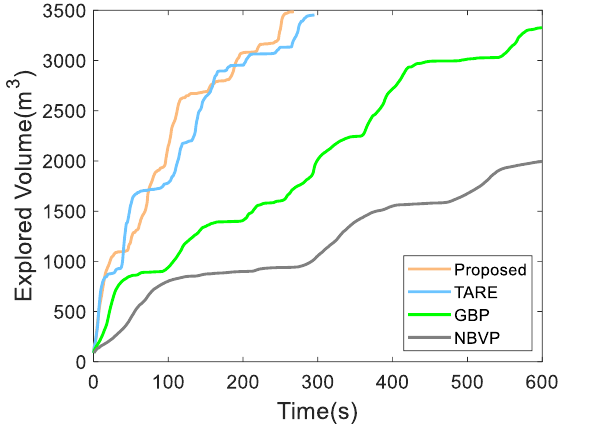}
}
\subfigure[]{
\includegraphics[width=0.21\textwidth]{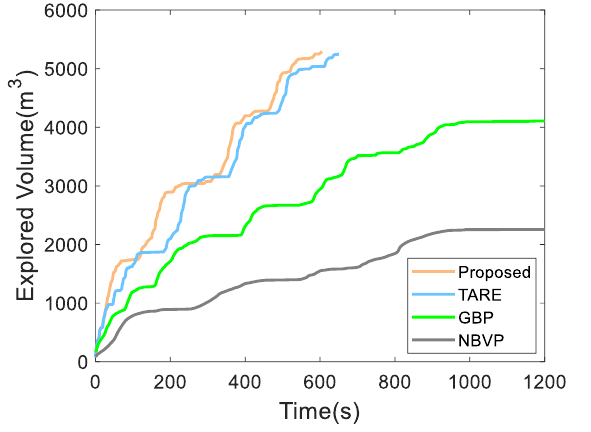}
}

\subfigure[]{
\includegraphics[width=0.21\textwidth]{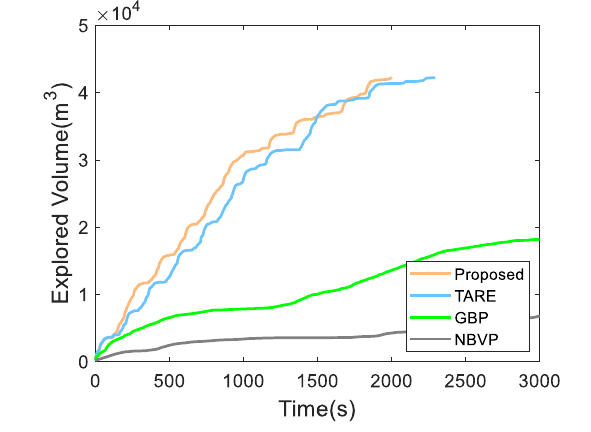}

}
\subfigure[]{
\includegraphics[width=0.21\textwidth]{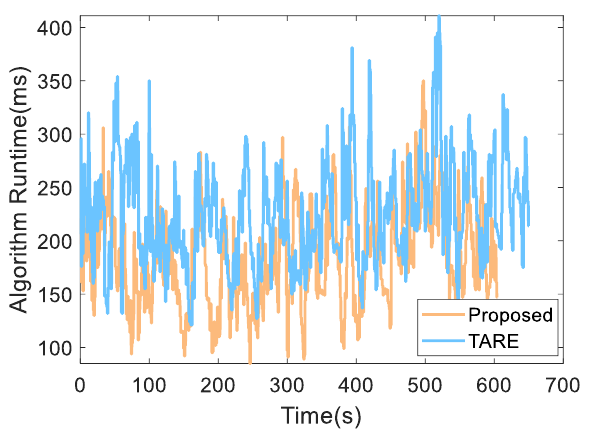}
}
\caption{Simulation results. (a), (b) and (c) correspond to Scene 1, Scene 2 and Scene 3 respectively. (d) is the result of algorithm runtime comparison of proposed method and TARE in Scene 2.}
\label{fig:diag}
\end{figure}

\begin{table}[!t]
\belowrulesep=0pt
\aboverulesep=0pt
\caption{Parameters of Our Method in Simulations}
\label{param}
\begin{center}
\begin{tabular}{c|c|c|c|c|c}
\toprule
  & \multicolumn{4}{c|}{\text{enclosed sub-region detection}} & \text{viewpoint refinement}\\ 
  \hline
  \text{parameter} & $r$ & $n$ &  $\lambda$ & $\alpha_{\text{max}}$ & $D_{\text{thr}}$ \\
  \hline
\text{value} & $1$m & $50$ & $4$ & $15^{\circ}$ & $7$m\\
  \bottomrule
\end{tabular}
\end{center}
\end{table}
\begin{table}[!t]
\belowrulesep=0pt
\aboverulesep=0pt
\caption{Results of Simulation Experiments in Three Scenes}
\label{result}
\begin{center}
\resizebox{0.48\textwidth}{!}
{\begin{tabular}{c|c|ccc|ccc}
\toprule
  \multirow{2}*{\textbf{Scene}} & \multirow{2}*{\textbf{Method}} & \multicolumn{3}{c|}{\textbf{Exploration time(s)}} & \multicolumn{3}{c}{\textbf{Movement distance(m)}}\\ 
  \cmidrule{3-8}
  & & \textbf{Avg} & \textbf{Max} &  \textbf{Min} & \textbf{Avg} & \textbf{Max} &  \textbf{Min}\\
  \midrule
  \multirow{4}{*}{Scene1} & NBVP & >1500 & - & - &  >573  & - & -\\
  & GBP & 815.5 & 923.2 & 779.1 & 646.4 & 687.1 & 601.9 \\
  & TARE & 295.3 & 314.2 & 284.5 & 550.7 & 568.1 & 535.3 \\
  & Proposed & \textbf{242.8} & 267.9 & 239.7 & \textbf{454.9}  & 482.1 & 441.6\\
  \cmidrule(lr){1-8}
  \multirow{4}{*}{Scene2}& NBVP & >2000 & - & - &  >659 & - & - \\
  & GBP & 1691.7 & 1779.6 & 1580.8 & 1758.9 & 1877.5 & 1660.8 \\
  & TARE & 650.3 & 689.5 & 641.5 & 1180.2 & 1221.4 & 1149.1 \\
  & Proposed & \textbf{582.7} & 621.8 & 566.9 & \textbf{1086.2} & 1159.1 & 1058.9 \\
  \cmidrule(lr){1-8}
  \multirow{4}{*}{Scene3}& NBVP & >4000 & - & - &  >1283 & - & - \\
  & GBP & >4000 & - & - &>1877 & - & - \\
  & TARE & 2293.3 & 2489.5 & 2100.5 & 4271.2 & 4303.4 & 3956.1 \\
  & Proposed & \textbf{2001.6} & 2290.1 & 1887.5 & \textbf{3765.2} & 3978.1 & 3630.9 \\
  \bottomrule
\end{tabular}}
\begin{tablenotes}
    \item NBVP\cite{bircher2016receding}. GBP\cite{dang2020graph}. TARE\cite{cao2021exploring}.
\end{tablenotes}
\end{center}
\end{table}

\section{Experiments and Results}

\begin{figure*}[htbp]
\centering
\subfigure[]{
\includegraphics[width=0.3\textwidth,  height=0.3\linewidth]{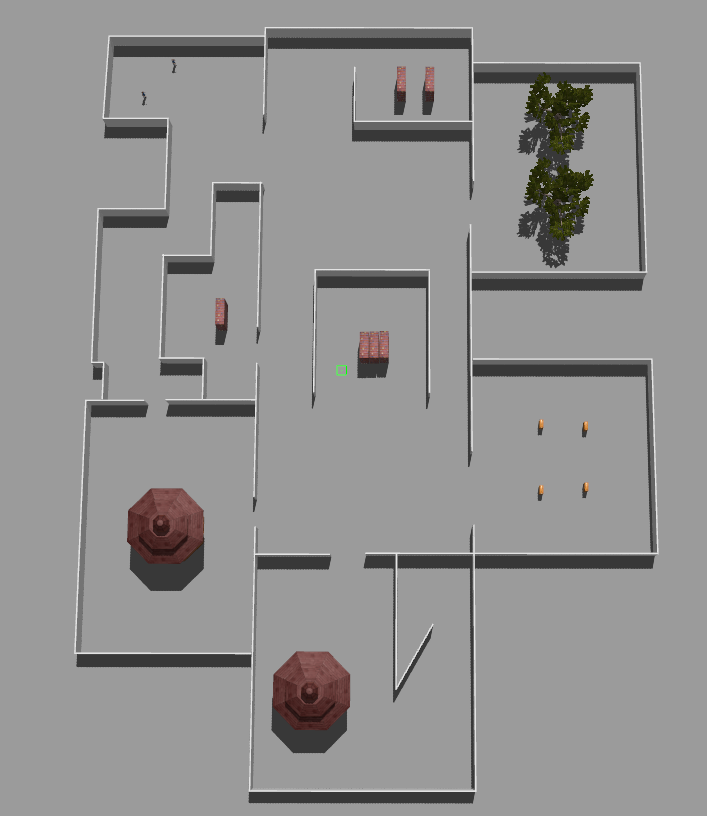}
}
\subfigure[]{
\includegraphics[width=0.3\textwidth,  height=0.3\linewidth]{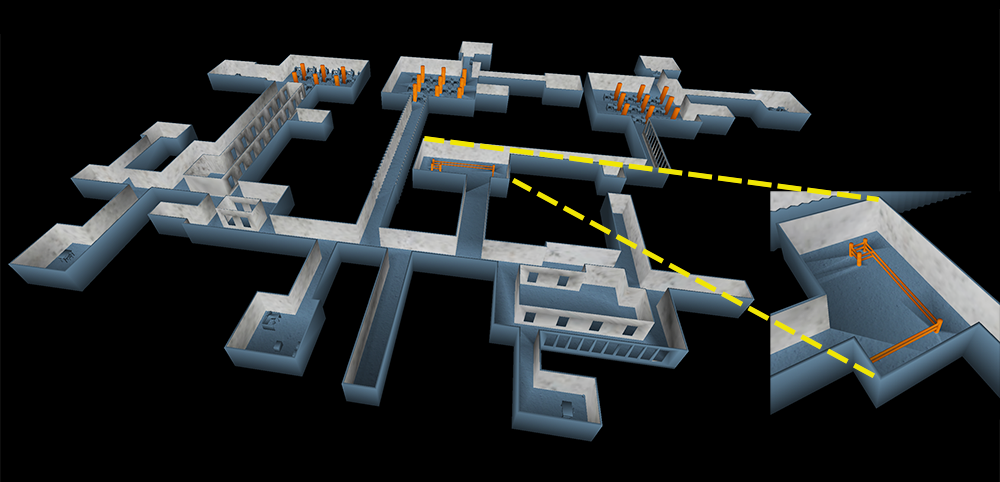}
}
\subfigure[]{
\includegraphics[width=0.3\textwidth,  height=0.3\linewidth]{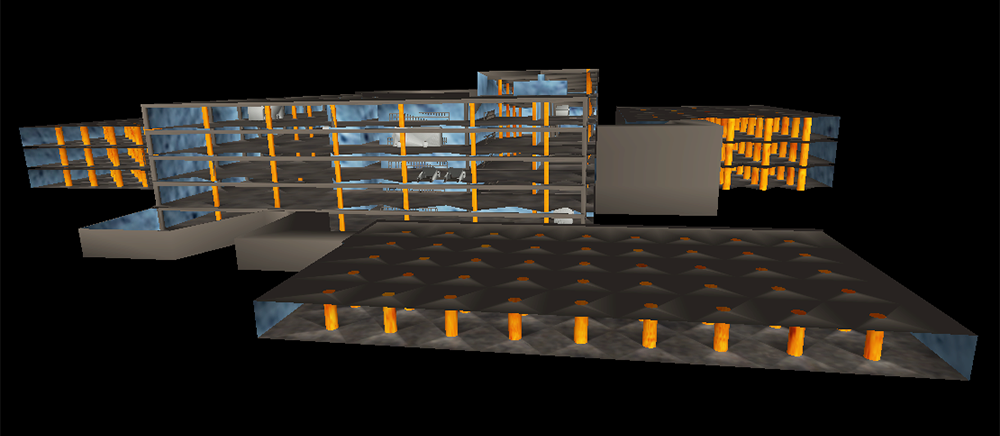}
}
\caption{The three simulation environments: (a) is a complex scenario that contains clutter structures and enclosed sub-regions. (b) is a large-scale indoor scenario. (c) is a multi-level garage scenario.}
\label{scene}
\end{figure*}

\begin{figure*}[htbp]
\centering
\subfigure[]{
\includegraphics[width=0.3\textwidth,  height=0.32\linewidth]{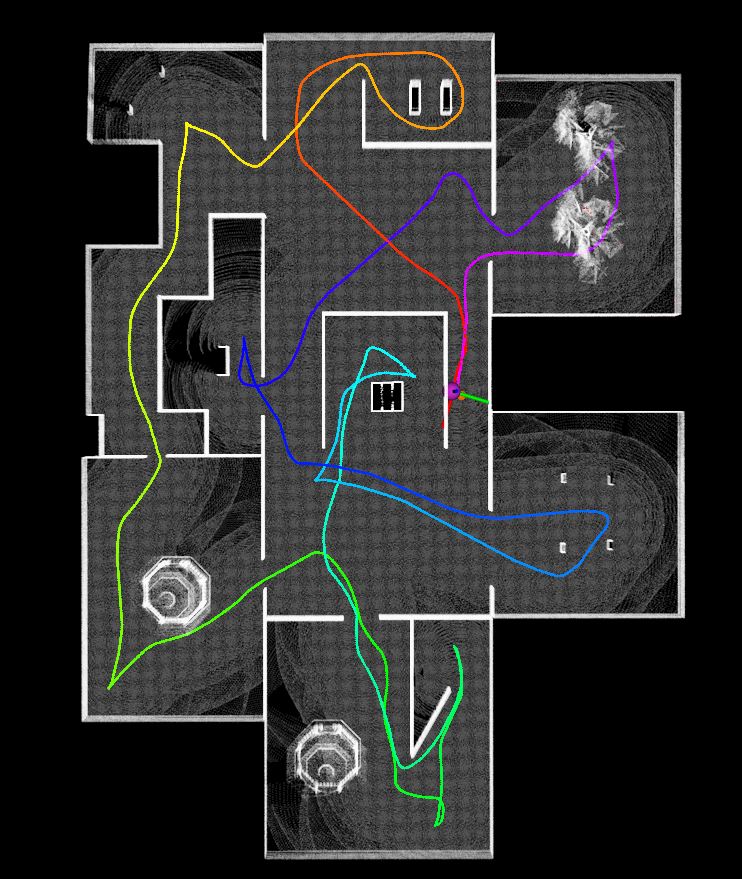}
}
\subfigure[]{
\includegraphics[width=0.3\textwidth,  height=0.32\linewidth]{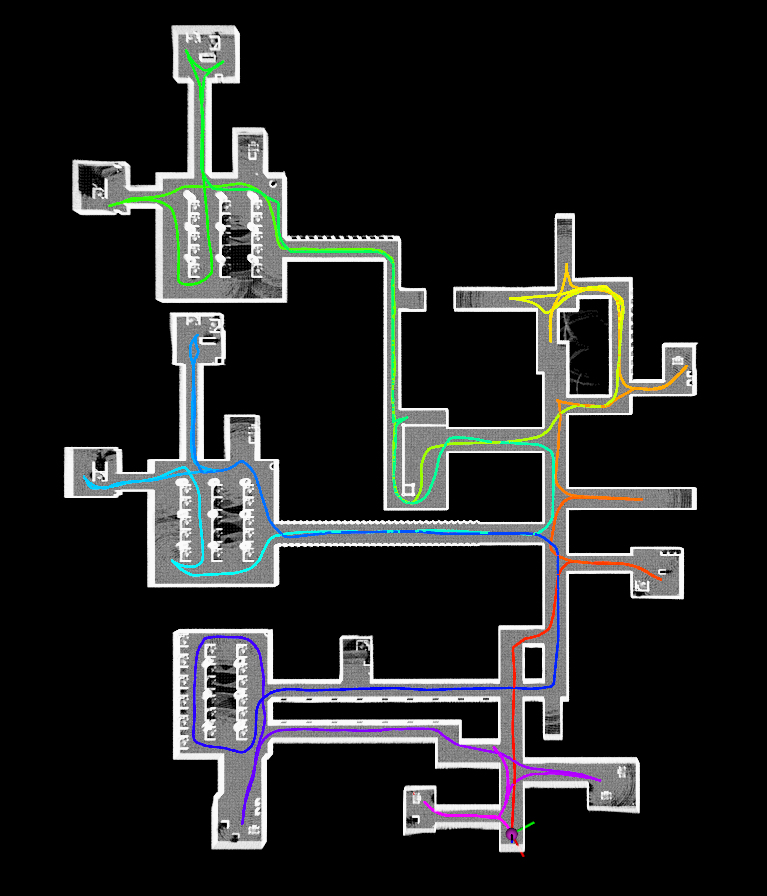}
}
\subfigure[]{
\includegraphics[width=0.3\textwidth,  height=0.32\linewidth]{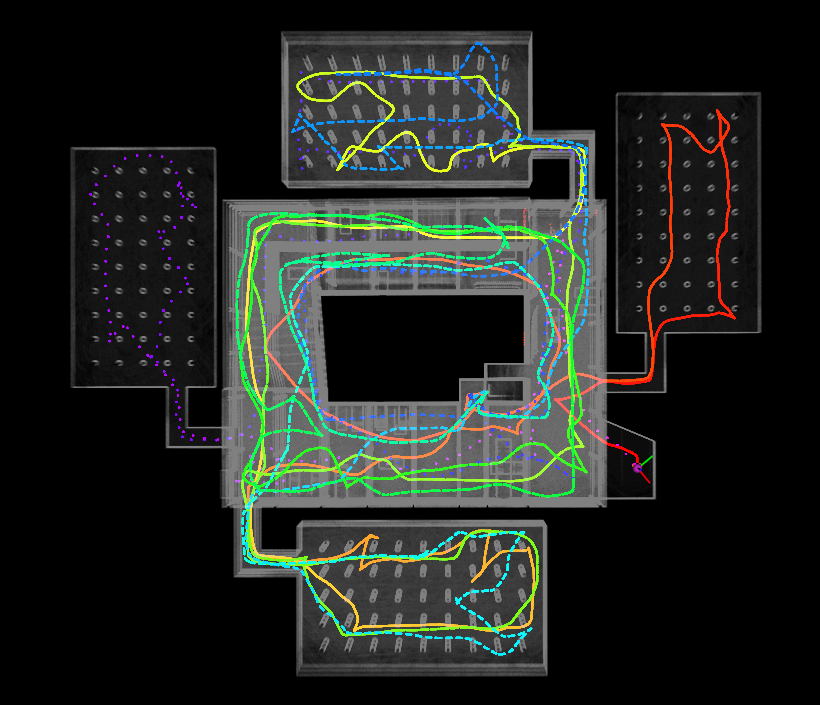}
}
\caption{Exploration trajectories (colored line) and reconstructed point cloud maps (white points) of the proposed method in three scenes. Videos of the experiments can be found at https://youtu.be/8a-ATr9pwhw}
\label{traj}
\end{figure*}

\subsection{Benchmark and Analysis}

The proposed method is tested in various environments to validate the efficiency by comparing with NBVP \cite{bircher2016receding} and GBP \cite{dang2020graph}, which are based on sampling strategy,  and TARE \cite{cao2021exploring}, which is based on frontier strategy. For all three methods, we adopt their open-source configuration. To compare the efficiency fairly, all simulations are run on a computer with Intel Core i5-12400F CPU and 16GB of RAM with Ubuntu 18.04 system. To prevent excessive time expenditure, we deem the exploration concluded when the GBP and NBVP algorithms are no longer exploring new areas or when the time elapsed exceeds twice the duration of our method.

As is shown in Fig. \ref{scene}, Scene 1 is constructed using Gazebo, Scene 2 and Scene 3 is provided by TARE. We set the size of the cubic region surrounding the robot for enclosed sub-region detection in Section \ref{III-B} to be 20m, 20m, and 5m in length, width, and height, respectively. In Equ. \ref{equ6}, the cost weights are set as $w_r=0.3$, $w_l=0.1$ and $w_d=0.2$. Other parameters are listed in Table \ref{param}. In both local and global planning, Lin-Kernighan-Helsgaun heuristic solver \cite{helsgaun2000effective} is employed to solve the ATSP. The sensor used in the simulation experiments is a Velodyne VLP-16 LiDAR with a FoV[$360^{\circ}$, $30^{\circ}$] and its max perception range is set to 15m. The max speed of the robot is set to 2m/s. Each method is tested ten times, with all methods starting from the same position in each scene to eliminate the impact of other factors on the experimental outcomes.

Table \ref{result} shows the results of the simulations. Exploration time and movement distance are utilized to measure the efficiency of different exploration approaches. To evaluate the performance of the four methods, we record the average values of these two metrics and show them in a diagram for a more intuitive representation as shown in Fig. \ref{fig:diag}(a), (b) and (c). Fig. \ref{fig:diag}(d) represents the comparative results of algorithm runtime between the proposed method and TARE. The results indicates that the proposed method  achieves a reduction in frontier detection time by a factor of 6 to 9 compared to TARE, decreasing from 55ms to 7ms. Concurrently, with the incorporation of an enclosed sub-region detection module, the overall runtime of our algorithm remains less than that of TARE.

\subsubsection{Scene 1: Complex Maze Scenario}
The scene range of Scene 1 is 90 $\times$ 60 $\times$ 3$\text{m}^\text{3}$. Scene 1 contains many enclosed sub-regions and cluttered structures. TARE complete the exploration in 295s and travels more than 500m. Due to many enclosed sub-regions in Scene 1, TARE expends a considerable amount of time and traverses repeated paths to complete the exploration of small regions. Our method completes the exploration in 242s and travels 454m, an 18\% faster performance compared to TARE. GBP completed the exploration in 815s. NBVP takes 1500s and does not completed the exploration. Due to the presence of many narrow passages in the scene, GBP and NBVP, being sampling-based exploration algorithms, exhibit significant randomness and therefore have difficulty navigating through narrow entrances, which is a significant factor for the low efficiency.

\begin{figure}[!t]
\centering
\includegraphics[width=0.28\textwidth]{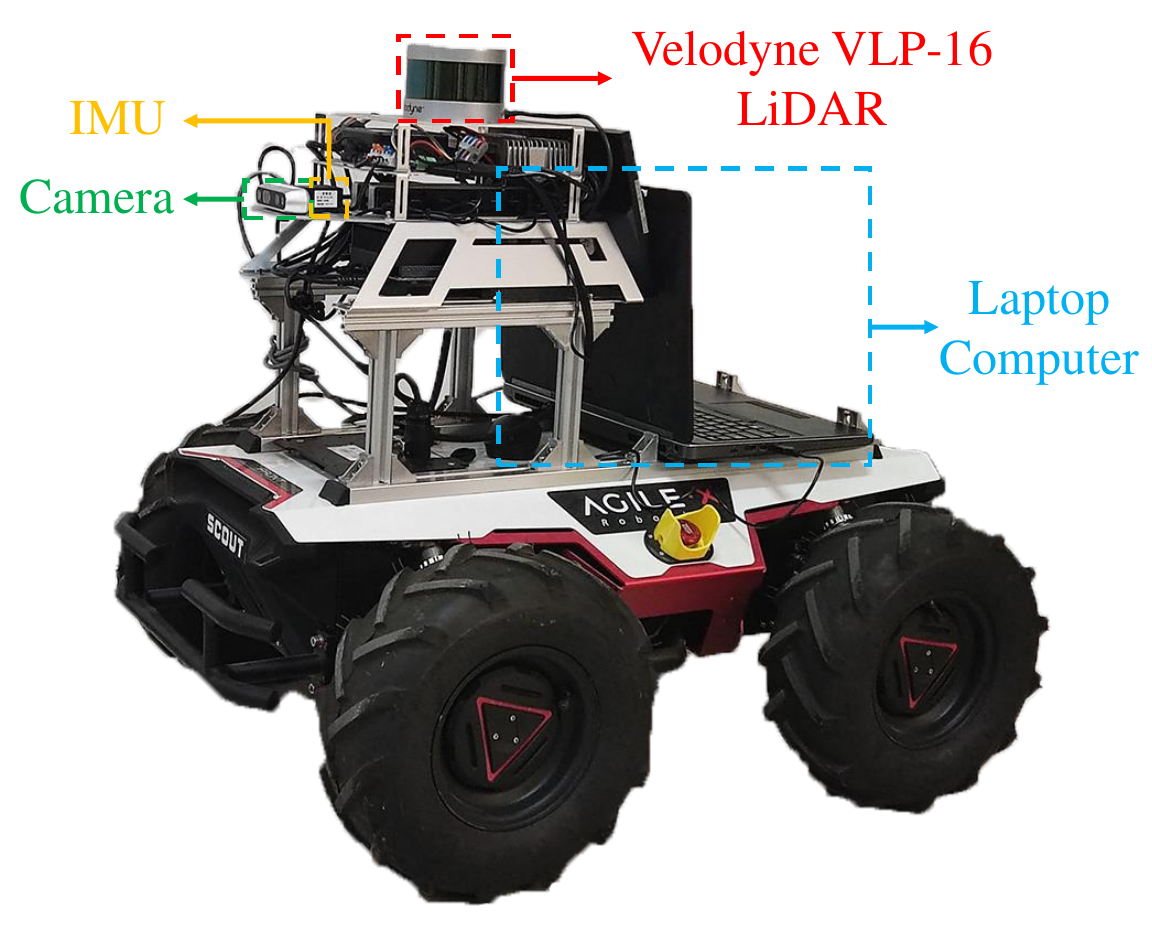}
\caption{The autonomous exploration platform.}
\label{real_vehicle}
\end{figure}

\begin{figure}[!t]
\centering
\subfigure[]{
\includegraphics[width=0.14\textwidth,  height=0.2\linewidth]{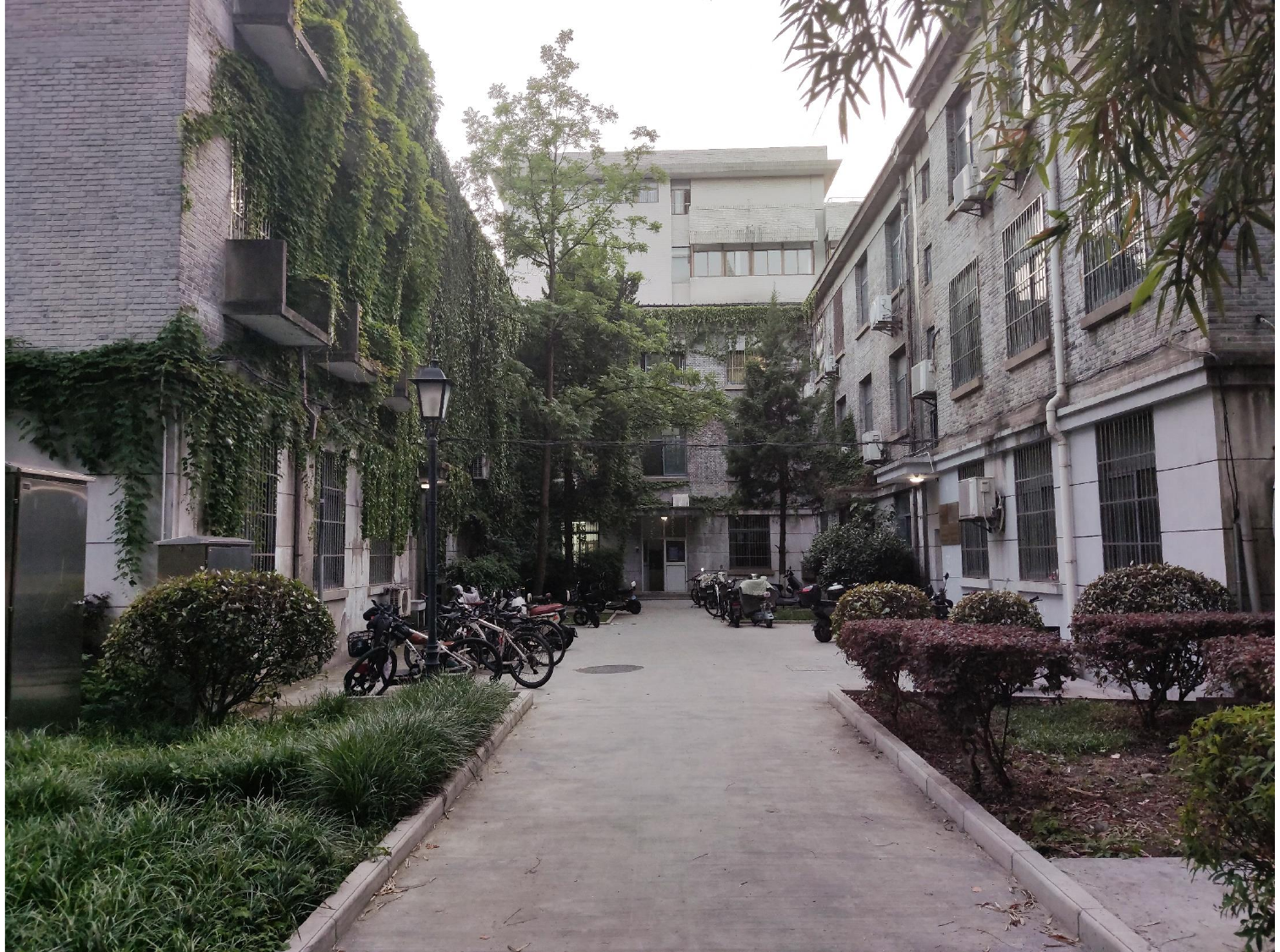}
}
\subfigure[]{
\includegraphics[width=0.14\textwidth,  height=0.2\linewidth]{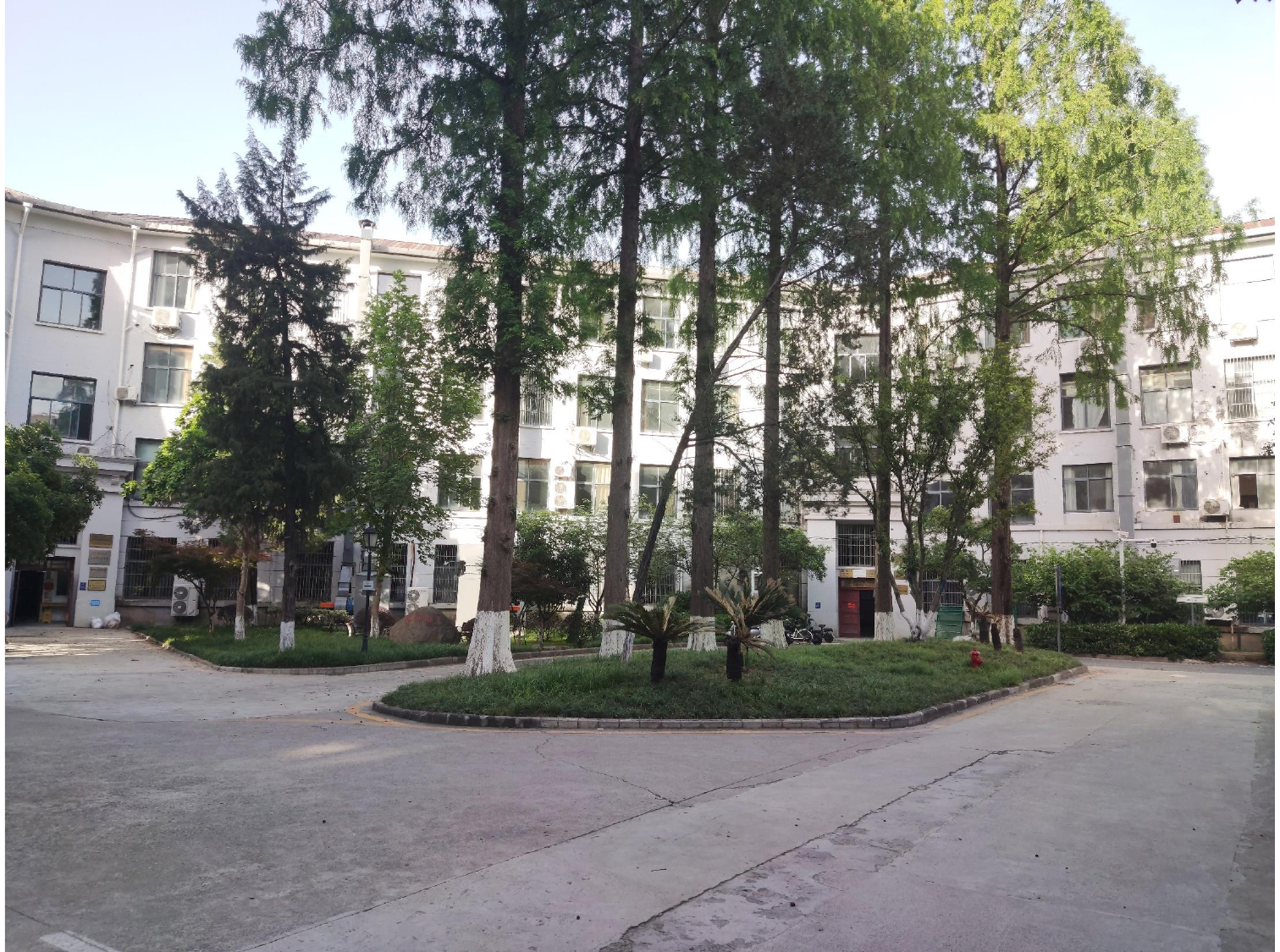}
}
\subfigure[]{
\includegraphics[width=0.14\textwidth,  height=0.2\linewidth]{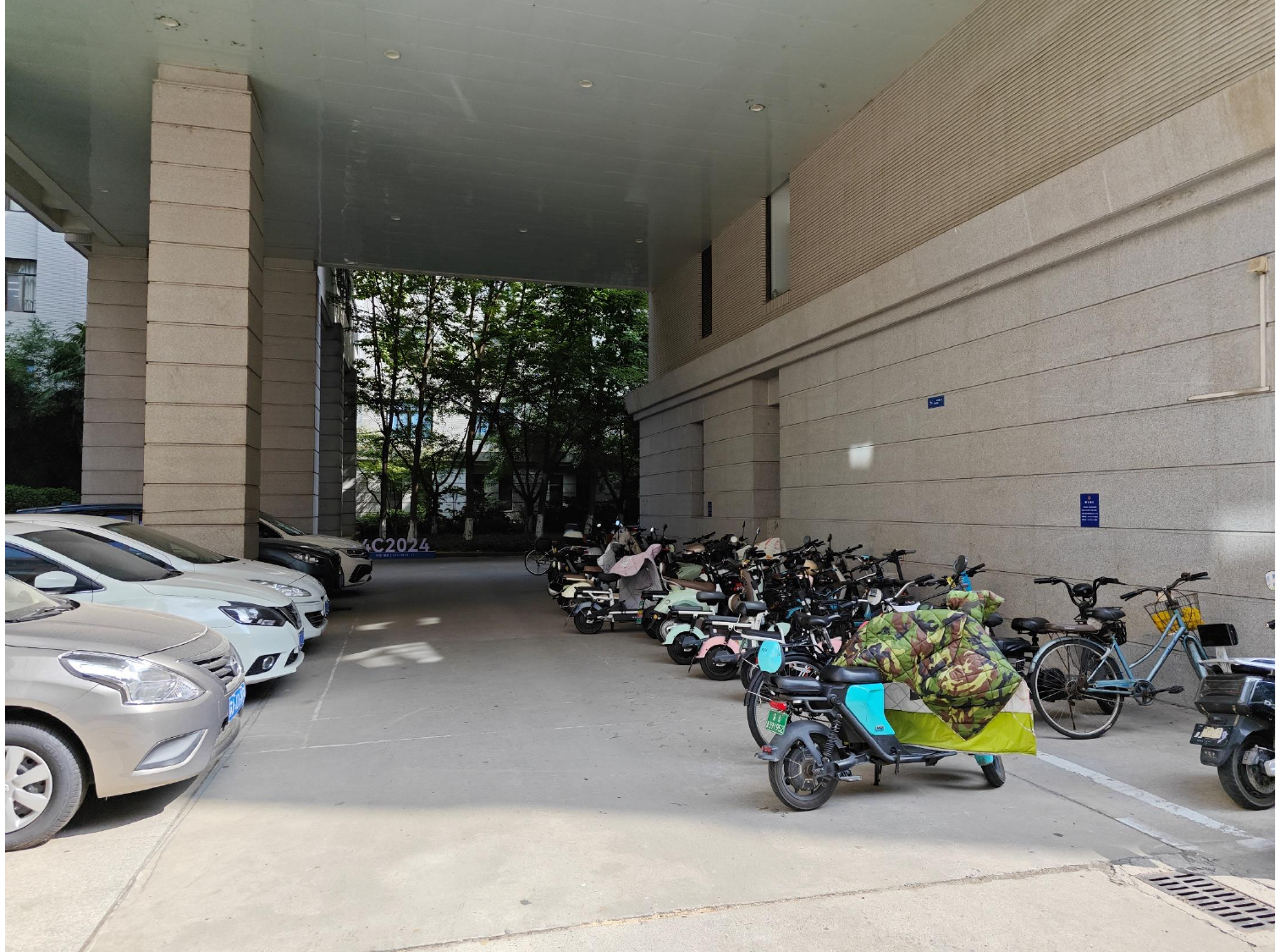}
}
\subfigure[]{
\includegraphics[width=0.43\textwidth, height=0.8\linewidth]{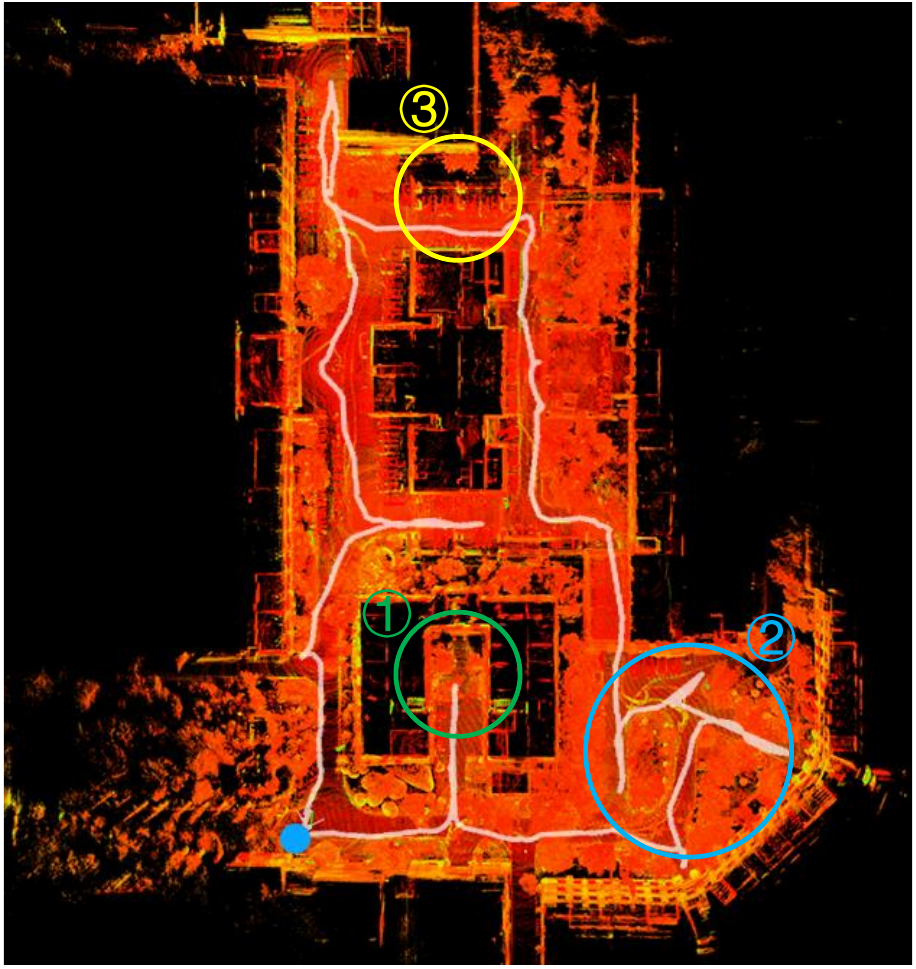}
}
\caption{The real-world experiment in outdoor environment. (a), (b) and (c) are the environments to be explored, which correspond to scenarios 1, 2 and 3 in (d), respectively. (d) is the reconstructed map of the exploration. The white line is the trajectory of the robot and the blue point is the starting and end position.}
\label{real_scene}
\end{figure}

\subsubsection{Scene 2: Cluttered and Large-scale Indoor Scenario}
The scene range of Scene 2 is 130 $\times$ 100 $\times$ 4$\text{m}^\text{3}$. Scene 2 is a large-scale and cluttered indoor scenario, which contains plenty of dead-ends and long galleries. As is shown in Table \ref{result}, TARE can complete the exploration in 650s, while our method takes 10\% less time than TARE. GBP completes the exploration three times slower than our method, while NBVP has not finished the exploration even after 2000s.

\subsubsection{Scene 3: Multi-level garage Scenario}
The scene range of Scene 3 is 100 $\times$ 100 $\times$ 15$\text{m}^\text{3}$. Scene3 is a challenging scenario for exploration task due to its context of multiple levels, complex layout with narrow passages and pillars. In the garage, robots are prone to omitting narrow regions, necessitating the traversal of redundant paths to accomplish the exploration of the entire environment. Our method complete the exploration in 2001s, while TARE spends 2293s. GBP and NBVP has not finished the exploration in 4000s, twice of the duration employed by our method.

\subsection{Real-World Experiments}

\begin{figure}[!t]
\centering
\includegraphics[width=0.4\textwidth]{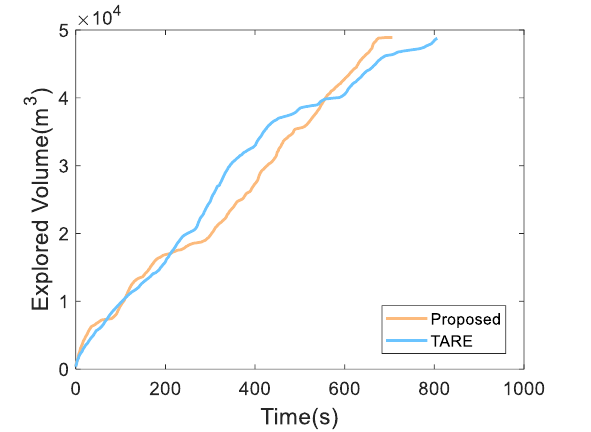}
\caption{Exploration result in real-world environment.}
\label{real_diag}
\end{figure}

In real-world environments, there are several challenges for autonomous exploration due to various uncertainties, such as pedestrians, vegetation and moving vehicles, that may influence the selection of the exploration path. To validate the effectiveness of our approach, we conducted experiments in an open school environment (Fig. \ref{real_scene}) using the vehicle platform shown in Fig. \ref{real_vehicle}. The vehicle is equipped with a Velodyne VLP-16 LiDAR and a MEMS-based IMU. Our algorithm runs on a laptop computer with Intel Core i7-7820HQ CPU and 32GB of RAM with Ubuntu 18.04 system. The LIO-SAM \cite{shan2020lio} is employed as the module of state estimation. The size of the cubic region surrounding the robot for enclosed sub-region detection in Section \ref{III-B} is set to be 20m, 20m, and 6m in length, width, and height, respectively. The max speed of the vehicle is set to 1.0m/s. The parameters of our method is set to $r=2$m, $\lambda=6$, and other parameters are the same as Table \ref{param}.

We conduct the real-world experiments in the outdoor environment as shown in Fig. \ref{real_scene}, where the explored region is bounded by a 200m $\times$ 120m rectangle. The results of the real-world experiment are shown in Fig. \ref{real_scene}(d) and Fig. \ref{real_diag}. The total volume explored is 48904$\text{m}^\text{3}$, the exploration time of the whole process is 706s, and the movement distance is 632m. The results of the real-world experiment show that our method can prioritize the exploration of enclosed sub-regions, thereby avoiding repeated paths and enhancing the efficiency of the exploration process.

\section{Conclusion}

In this letter, we propose an efficient exploration planner for complex and unknown environments. A fast frontier updating mechanism is designed, which reduces the requirement of computational resources significantly. To improve the efficiency of exploration, we propose a concept of enclosed sub-region and devise a viewpoint evaluation system to increase the visiting priority of viewpoints in enclosed sub-regions, reducing repeated paths without prior knowledge of the scene. In addition, we raise a viewpoint refinement approach to generate smoother paths. The benchmark analysis shows that our method can improve the efficiency of exploration by 10\%$\sim$20\% compared to the state-of-the-art method.
Both simulation and real-world experiments validate the high efficiency of our approach.

\addtolength{\textheight}{-12cm}   








\bibliographystyle{IEEEtran}
\bibliography{IEEEabrv,reference}

\end{document}